\documentclass[11pt,letterpaper]{article}
\usepackage{emnlp2017}
\usepackage{times}
\usepackage{latexsym}

\usepackage{url}
\usepackage{amssymb}
\usepackage{amsmath}
\usepackage{stmaryrd}
\usepackage{array}
\usepackage{graphicx}
\usepackage{cases}
\usepackage{watermark}
\usepackage{pbox}
\usepackage{multirow}
\usepackage{soul}
\setulcolor{red}
\usepackage{enumitem}
\usepackage{xspace}
\usepackage{makecell}
\usepackage[draft]{todonotes} 
\presetkeys{todonotes}{fancyline, color=pink, bordercolor=red}{}
\usepackage[ruled,noline,linesnumbered]{algorithm2e}
\usepackage[small]{caption}
\usepackage{lipsum}
\usepackage{mathtools}
\usepackage{cuted}

\usepackage{xcolor}

\newcommand{\metric}{GLEU\xspace}
\newcommand{\proposed}{NRL\xspace}

\newcommand{\encdec}{enc-dec\xspace}

\newcommand{\mrt}{MRT\xspace}

\newcommand{\mle}{MLE\xspace}
\newcommand{\gec}{GEC\xspace}
\newcommand{\cambhybrid}{CAMB14\xspace}
\newcommand{\cambnmt}{CAMB16\xspace}
\newcommand{\nus}{NUS\xspace}
\newcommand{\amu}{AMU\xspace}

\newcommand{\trainingdata}{(X,Y)}
\newcommand{\mrtdelta}{\Delta(\hat{y},y)}
\newcommand{\nrlreward}{r(\hat{y},y)}
\newcommand{\sampledata}{\hat{y} \in S(x)}
\newcommand{\sampleprimedata}{\hat{y}' \in S(x)}

\newcommand{\mrtp}{p(\hat{y}| x;\theta)}
\newcommand{\mrtpsimple}{p(\hat{y})}
\newcommand{\mrtpprime}{p(\hat{y}'| x;\theta)}
\newcommand{\mrtpprimesimple}{p(\hat{y}')}
\newcommand{\mrtpsum}{\sum_{\sampleprimedata} \mrtpprime}
\newcommand{\mrtq}{q(\hat{y}|x;\theta,\alpha)}
\newcommand{\mrtqsimple}{q(\hat{y})}
\newcommand{\softmax}{\frac{\mrtp^\alpha}{\mrtpsum^\alpha}}

\newcommand{\mrtpsumsimple}{\sum_{\hat{y}'} \mrtpprimesimple}
\newcommand{\softmaxnumerator}{\mrtpsimple^\alpha}
\newcommand{\softmaxdenominator}{\mrtpsumsimple^\alpha}

\newcommand{\pderivrtilde}{\frac {\partial \tilde{R}(\theta)}{\partial \theta}}
\newcommand{\pderivl}{\frac {\partial L(\theta)}{\partial \theta}}

\newcommand{\pderivj}{\frac {\partial J(\theta)}{\partial \theta}}
\newcommand{\pderivtheta}{\frac {\partial}{\partial \theta}}
\newcommand{\pderivp}{\frac {\partial}{\partial \mrtpsimple}}

\emnlpfinalcopy

\def\emnlppaperid{***}

\newcommand{\clsp}{\ensuremath{{}^\dagger}}
\newcommand{\hltcoe}{\ensuremath{{}^\ddagger}}

\title{Grammatical Error Correction with Neural Reinforcement Learning}

\author{
  Keisuke Sakaguchi \clsp
  \and Matt Post\hltcoe
  \and Benjamin Van Durme\clsp \hltcoe \\
 \clsp Center for Language and Speech Processing, Johns Hopkins University \\
 \hltcoe Human Language Technology Center of Excellence, Johns Hopkins University \\
 {\tt \{keisuke,post,vandurme\}@cs.jhu.edu}
}
\date{}

\begin{document}
\setlength{\abovedisplayskip}{3.0pt} 
\setlength{\belowdisplayskip}{3.0pt} 

\maketitle

\begin{abstract}
We propose a neural encoder-decoder model with reinforcement learning (\proposed) for grammatical error correction (GEC).
Unlike conventional maximum likelihood estimation (\mle), the model directly optimizes towards an objective that considers a {\em sentence-level}, task-specific evaluation metric, avoiding the exposure bias issue in \mle.
We demonstrate that \proposed outperforms \mle both in human and automated evaluation metrics, achieving the state-of-the-art on a fluency-oriented GEC corpus. 
\end{abstract}

\section{Introduction}
Research in automated Grammatical Error Correction (\gec) has expanded  from token-level, closed class corrections (e.g., determiners, prepositions, verb forms) to phrase-level, open class issues that consider fluency (e.g., content word choice, idiomatic collocation, word order, etc.).

The expanded goals of GEC have led to new proposed models deriving from techniques in data-driven machine translation, including phrase-based MT (PBMT) \cite{felice-EtAl:2014:W14-17,chollampatt-hoang-ng:2016:EMNLP2016,junczysdowmunt-grundkiewicz:2016:EMNLP2016} and neural encoder-decoder models (\encdec) \cite{yuan-briscoe:2016:N16-1}.
\newcite{napoles-sakaguchi-tetreault:2017:EACLshort} recently showed that a neural \encdec can outperform PBMT on a fluency-oriented GEC data and metric.

We investigate training methodologies in the neural \encdec for GEC.
To train the neural \encdec models, maximum likelihood estimation (\mle) has been used, where the objective is to maximize the (log) likelihood of the parameters for a given training data.

\newlength{\textfloatsepsave}
\setlength{\textfloatsepsave}{\textfloatsep} \setlength{\textfloatsep}{0pt}
\begin{algorithm}[t]
\DontPrintSemicolon
\SetAlgoNoEnd
\KwIn{Pairs of source ($X$) and target ($Y$)}
\KwOut{Model parameter $\hat{\theta}$}
initialize($\hat{\theta}$)\\
\For{$(x,y) \in (X,Y)$}{
  ${\small (\hat{y}_1, ... \hat{y}_k), (p(\hat{y}_1), ... p(\hat{y}_k)) = \text{sample}(x, k, \hat{\theta})}$\\
  $p(\hat{y})$ = normalize($p(\hat{y})$) \\
  $\bar{r}(\hat{y}) = 0$ \tcp*[r]{\small expected reward} 
  \For{$\hat{y}_i \in \hat{y}$}{
    $\bar{r}(\hat{y}) += p(\hat{y}_i)\cdot \text{score}(\hat{y}_i, y)$ \\
  }
  backprop($\hat{\theta}, \bar{r}$) \tcp*[r]{\small policy gradient $\frac{\partial}{\partial \hat{\theta}}$}
}
\KwRet{$\hat{\theta}$}\\
\caption{Reinforcement learning for neural encoder-decoder model.}
\label{alg:nrl}
\end{algorithm}
As \newcite{2015arXiv151106732R} indicates, however, \mle has drawbacks.
The \mle objective is based on {\em word-level} accuracy against the reference, and the model is not exposed to the predicted output during training (exposure bias).
This becomes problematic, because once the model fails to predict a correct word, it falls off the right track and does not come back to it easily.

To address the issues, we employ a neural \encdec GEC model with a reinforcement learning approach in which we directly optimize the model toward our final objective (i.e., evaluation metric).
The objective of the neural reinforcement learning model (\proposed) is to maximize the expected reward on the training data.
The model updates the parameters through back-propagation according to the reward from predicted outputs. 
The high-level description of the training procedure is shown in Algorithm \ref{alg:nrl},
and more details are explained in \S\ref{sec:model}.
To our knowledge, this is the first attempt to employ reinforcement learning for directly optimizing the \encdec model for \gec task.

We run GEC experiments on a fluency-oriented GEC corpus (\S\ref{sec:experiment}), demonstrating that \proposed outperforms the \mle baseline both in human and automated evaluation metrics.

\section{Model and Optimization}
\vspace{-2mm}
\label{sec:model}
\setlength{\textfloatsep}{\textfloatsepsave}
We use the {\em attentional neural \encdec} model \cite{2014arXiv1409.0473B} as a basis for both \proposed and \mle.
The model takes (possibly ungrammatical) source sentences $x \in X$ as an input, and predicts grammatical and fluent output sentences $y \in Y$ according to the model parameter $\theta$.
The model consists of two sub-modules, {\em encoder} and {\em decoder}. 
The encoder transforms $x$ into a sequence of vector representations (hidden states) using a bidirectional gated recurrent neural network (GRU) \cite{2014arXiv1412.3555C}.
The decoder predicts a word $y_t$ at a time, using previous token $y_{t-1}$ and linear combination of encoder information as attention.
%

\vspace{-1mm}
\subsection{Maximum Likelihood Estimation}
\vspace{-1mm}
\label{sec:mle}
Maximum Likelihood Estimation training (\mle) is a standard optimization method for \encdec models.
In \mle, the objective is to maximize the log likelihood of the correct sequence for a given sequence for the entire training data.
\begin{eqnarray}
L(\theta) = \sum_{\langle X,Y \rangle} \sum_{t=1}^{T}\log p(y_{t}|x, y_1^{t-1};\theta)
\end{eqnarray}
The gradient of $L(\theta)$ is as follows:
\begin{eqnarray}
\pderivl = \sum_{\langle X,Y \rangle} \sum_{t=1}^{T} \frac{\nabla p(y_t|x, y_1^{t-1};\theta)}{p(y_t|x,y_1^{t-1};\theta)}
\end{eqnarray}

One drawback of \mle is the {\em exposure bias} \cite{2015arXiv151106732R}. 
The decoder predicts a word conditioned on the correct word sequence ($y_{1}^{t-1}$) during training, whereas it does with the predicted word sequence ($\hat{y}_{1}^{t-1}$) at test time. 
Namely, the model is not exposed to the predicted words in training time.
This is problematic, because once the model fails to predict a correct word at test time, it falls off the right track and does not come back to it easily.
Furthermore, in most sentence generation tasks, the \mle objective does not necessarily correlate with our final evaluation metrics, such as BLEU \cite{papineni-EtAl:2002:ACL} in machine translation and ROUGE \cite{lin:2004:ACLsummarization} in summarization.
This is because \mle optimizes word level predictions at each time step instead of evaluating sentences as a whole.

\gec is no exception.
It depends on sentence-level evaluation that considers grammaticality and fluency.
For this purpose, it is natural to use \metric \cite{napoles-EtAl:2015:ACL-IJCNLP}, which has been used as a fluency-oriented GEC metric. 
We explain more details of this metric in \S\ref{sec:gleu}.


\vspace{-1mm}
\subsection{Neural Reinforcement Learning}
\vspace{-1mm}
\setlength{\abovedisplayskip}{4.0pt} 
\setlength{\belowdisplayskip}{4.0pt} 
To address the issues in \mle, we directly optimize the neural \encdec model toward our final objective for GEC using reinforcement learning.
In reinforcement learning, {\em agents} aim to maximize expected {\em rewards} by taking {\em actions} and updating the {\em policy} under a given {\em state}.
In the neural \encdec model, we treat the \encdec as an agent which predicts a word from a fixed vocabulary at each time step (the action), given the hidden states of the neural \encdec representation.
The key difference from \mle is that the reward is not restricted to token-level accuracy. 
Namely, any arbitrary metric is applicable as the reward.\footnote{The reward is given at the end of the decoder output (i.e., delayed reward).}

Since we use \metric as the final evaluation metric, the objective of \proposed is to maximize the expected \metric by learning the model parameter. 
\begin{align}
\label{eq:j}
J(\theta) &= \mathbb{E}[\nrlreward] \nonumber \\
    &= \sum_{\sampledata} \mrtp \nrlreward
\end{align}
where $S(x)$ is a sampling function that produces $k$ samples $\hat{y}_1, ... \hat{y}_k$, $\mrtp$ is a probability of the output sentence, and $\nrlreward$ is the reward for $\hat{y}_k$ given a reference set $y$.
As described in Algorithm \ref{alg:nrl}, given a pair of source sentence and the reference $(x, y)$, \proposed takes $k$ sample outputs ($\hat{y}_1$, ... $\hat{y}_k$) and their probabilities ($p(\hat{y}_1)$, ... $p(\hat{y}_k)$).
Then, the expected reward is computed by multiplying the probability and metric score for each sample $\hat{y}_i$.

In the \encdec, the parameters $\theta$ are updated through back-propagation and the number of parameter updates is determined by the partial derivative of $J(\theta)$, called the {\em policy gradient} \cite{williams1992simple,sutton1999policy} in reinforcement learning:
\begin{align}
\label{eq:partialj}
\pderivj = \alpha \mathbb{E} \left[\nabla \log \mrtpsimple \{\nrlreward - b \}  \right]
\end{align}
where $\alpha$ is a learning rate and $b$ is an arbitrary baseline reward to reduce the variance.
The sample mean reward is often used for $b$ \cite{williams1992simple}, and we follow it in \proposed.

It is reasonable to compare \proposed to minimum risk training (\mrt) \cite{shen-EtAl:2016:P16-1}.
In fact, \proposed with a {\em negative expected reward} can be regarded as \mrt.
The gradient of \mrt objective is a special case of {\em policy gradient} in \proposed.
We show mathematical details about the relevance between \proposed and \mrt in the supplemental material.

\subsection{Reward in Grammatical Error Correction}
\label{sec:gleu}
To capture fluency as well as grammaticality in evaluation on such references, we use \metric as the reward.
\metric has been shown to be more strongly preferred than other GEC metrics by native speakers \cite{TACL800}. 
Similar to BLEU in machine translation, \metric computes $n$-gram precision between the system hypothesis ($H$) and the reference ($R$).
In \metric, however, $n$-grams in source ($S$) are also considered. The precision is penalized when the $n$-gram in $H$ overlaps with the source and not with the reference.
Formally, 
\begin{align}
\text{\metric} &= \text{BP}\cdot \exp \left( \sum_{n=1}^4 \frac{1}{n} \log p'_n \right) \nonumber \\
p'_n &= \frac{ N(H,R) - \left[ N(H,S)-N(H,S,R) \right] }{N(H)} \nonumber
\end{align}
where  $N(A,B,C,...)$ is the number of overlapped $n$-grams among the sets, and BP is the same {\em brevity penalty} as in BLEU.

\section{Experiments}
\label{sec:experiment}
\vspace{-1mm}
\begin{table}[t]
\small
\centering
\begin{tabular}{l|c|c|c}
\hline
       &     & {\bf mean chars} & {\bf \# sents.} \\
{\bf Corpus} & {\bf \# sents.} & {\bf per sent.}  & {\bf edited}   \\ \hline\hline
NUCLE  & 57k & 115             & 38\%          \\
FCE    & 34k & 74              & 62\%          \\
Lang-8 & 1M  & 56              & 35\%          \\ \hline
\end{tabular}
\caption{Statistics of training corpora}
\label{tab:corpora}
\vspace{-4mm}
\end{table}

\vspace{-1mm}
\paragraph{Data} For training the models (\mle and \proposed), we use the following corpora: 
the NUS Corpus of Learner English (NUCLE) \cite{dahlmeier-ng-wu:2013:BEA8}, 
the Cambridge Learner Corpus First Certificate English (FCE) \cite{yannakoudakis-briscoe-medlock:2011:ACL-HLT2011}, and
the Lang-8 Corpus of learner English \cite{tajiri-komachi-matsumoto:2012:ACL2012short}. 
The basic statistics are shown in Table \ref{tab:corpora}.\footnote{All the datasets are publicly available, for purposes of reproducibility.}
We exclude some unreasonable edits (comments by editors, incomplete sentences such as URLs, etc.) using regular expressions and setting a maximum token edit distance within 50\% of the original length.
We also ignore sentences that are longer than 50 tokens or sentences where more than 5\% of tokens are out-of-vocabulary (the vocabulary size is 35k).
In total, we use 720k pairs of sentences for training (21k from NUCLE, 32k from FCE, and 667k from Lang-8).
Spelling errors are corrected in preprocessing with Enchant.\footnote{\url{https://github.com/AbiWord/enchant}}

\vspace{-2mm}
\paragraph{Hyperparameters in \encdec}
For both \mle and \proposed, we set the vocabulary size to be 35k for both source and target.
Words are represented by a vector with 512 dimensions.
Maximum output token length is 50.
The size of hidden layer units is 1,000.
Gradients are clipped at 1, and beam size during decoding is 5.
We regularize the GRU layer with a dropout probability of 0.2.

For \mle we use mini-batches of size 40, and the ADAM optimizer \cite{kingma2014adam} with a learning rate of $10^{-4}$.
We train the \encdec with \mle for 900k updates,  selecting the best model according to the development set evaluation.

For \proposed we set the sample size to be 20. 
We use the SGD optimizer with a learning rate of $10^{-4}$. 
For the {\em baseline reward}, we use average of sampled reward following \newcite{williams1992simple}.
The sentence GLEU score is used as the reward $\nrlreward$.
Following a similar (but not the same) strategy of the Mixed Incremental Cross-Entropy Reinforce (MIXER) algorithm \cite{2015arXiv151106732R}, we initialize the model by \mle for 600k updates, followed by another 600k updates using \proposed, and select the best model according to the development set evaluation.
Our \proposed is implemented by extending the Nematus toolkit \cite{sennrich-EtAl:2017:EACLDemo}.

\begin{table}[t]
\small
\centering
\begin{tabular}{l|l|l}
\hline
Models      & Methods          & \# sents (corpora)    \\ \hline \hline
\cambhybrid & Hybrid           & 155k \\
            & (rule + PBMT)    & (NUCLE, FCE, in-house) \\ \hline
\amu        & PBMT +           & 2.3M   \\
            & GEC-feat.        & (NUCLE, Lang8) \\ \hline
\nus        & PBMT +           & 2.1M \\
            & Neural feat.     & (NUCLE, Lang8)  \\ \hline
\cambnmt    & \encdec (\mle) + & 1.96M \\
            & unk alignment    & (non-public CLC)   \\ \hline \hline
\mle/\proposed & \encdec       & 720k \\ 
            & (\mle/\proposed) & (NUCLE, Lang8, FCE) \\ \hline
\end{tabular}
\caption{Summary of baselines, \mle and \proposed models.}
\label{tab:models}
\vspace{-4mm}
\end{table}  

 
\begin{table*}[t]
\fontsize{9}{11}\selectfont
\centering
\begin{tabular}{l|l}
\hline
Orig. & but found that successful people use the people money and use there idea for a way to success .\\ \hline
Ref.  & But it was found that successful people use other people 's money and use their ideas as a way to success .\\ \hline \hline
MLE   & But found that successful people use the people money and use it for a way to success .\\ \hline
NRL   & But found that successful people use the people 's money and use their idea for a way to success .\\ \hline \hline
Orig. & Fish firming uses the lots of special products such as fish meal .\\ \hline
Ref.  & Fish firming uses a lot of special products such as fish meal .\\ \hline \hline
MLE   & Fish contains a lot of special products such as fish meals .\\ \hline
NRL   & Fish shops use the lots of special products such as fish meal .\\ \hline
\end{tabular}
\caption{Example outputs by \mle and \proposed}
\label{tab:examples}
\vspace{-2mm}
\end{table*}
  
\vspace{-1mm}
\paragraph{Baselines}
In addition to our \mle baseline, we compare four leading GEC systems. 
All the systems are based on SMT, but they take different approaches.
The first model, proposed by \newcite{felice-EtAl:2014:W14-17}, uses a combination of a rule-based system and PBMT with language model reranking (referring as \cambhybrid).
\newcite{junczysdowmunt-grundkiewicz:2016:EMNLP2016} proposed a PBMT model that incorporates linguistic and GEC-oriented sparse features (\amu).
Another PBMT model, proposed by \newcite{chollampatt-hoang-ng:2016:EMNLP2016}, is integrated with neural contextual features (\nus).
Finally, \newcite{yuan-briscoe:2016:N16-1} proposed a neural \encdec model with \mle training (\cambnmt). 
This model is similar to our \mle model, but \cambnmt additionally trains an unsupervised alignment model to handle spelling errors as well as unknown words, and it uses 1.96M sentence pairs extracted from the non-public Cambridge Learner Corpus (CLC).
The summary of baselines is shown in Table \ref{tab:models}.\footnote{The four baselines are not tuned toward the same dev set as \mle and \proposed. Also, they use different training set (Table \ref{tab:models}). We compare them just for reference.}

\vspace{-1mm}
\paragraph{Evaluation}
For evaluation, we use the JFLEG corpus \cite{napoles-sakaguchi-tetreault:2017:EACLshort}, which consists of 1501 sentences (754: dev, 747: test) with four fluency-oriented references per sentence.

Regarding the evaluation metric, in addition to the automated metric (\metric), we run a human evaluation using Amazon Mechanical Turk (MTurk).
We randomly select 200 sentences each from the dev and test set.
For each sentence, two turkers are repeatedly asked to rank five systems randomly selected from all eight: the four baseline models, \mle, \proposed, one randomly selected human correction, and the original sentence.
We infer the evaluation scores by efficiently comparing pairwise rankings with the TrueSkill algorithm \cite{HerbrichMG06,sakaguchi-post-vandurme:2014:W14-33}.




\vspace{-1mm}
\paragraph{Results}
Table \ref{tab:results} shows the human evaluation by TrueSkill and automated metric (\metric).
In both dev and test set, \proposed outperforms \mle and other baselines in both the human and automatic evaluations.
Human evaluation and GLEU scores correlate highly, corroborating the reliability of \metric.
With respect to inter-annotator agreement, Spearman's rank correlation between Turkers is 55.6 for the dev set and 49.2 for the test set.
The correlations are sufficiently high to show the agreement between Turkers, considering the low chance level (i.e., ranking five randomly selected systems consistently between two Turkers).

\begin{table}[t]
\small
\centering
\begin{tabular}{l|c|c|c|c}
\hline
   & \multicolumn{2}{c|}{dev set} & \multicolumn{2}{c}{test set} \\ \hline
Models      & Human & GLEU  & Human  & GLEU \\ \hline  \hline
Original    & -1.072& 38.21 & -0.760 & 40.54\\ \hline
\amu        & -0.405& 41.74 & -0.168 & 44.85 \\
\cambhybrid & -0.160& 42.81 & -0.225 & 46.04 \\
\nus        & -0.131& 46.27 & -0.249 & 50.13 \\
\cambnmt    & -0.117& 47.20 & -0.164 & 52.05 \\ \hline
\mle        & -0.052& 48.24 & -0.110 & 52.75 \\
\proposed   & 0.169 & {\bf 49.82} & 0.111 & {\bf 53.98} \\ \hline \hline
Reference   & 1.769 & 55.26 & 1.565  & 62.37 \\ \hline
\end{tabular}
\caption{Human (TrueSkill) and GLEU evaluation of system outputs on the development and test set.}
\label{tab:results}
\vspace{-2mm}
\end{table}

\vspace{-1mm}
\paragraph{Analysis}
Table \ref{tab:examples} presents example outputs. 
In the first example, both \mle and \proposed successfully corrected the homophone error ({\em there vs. their}), but \mle changed the meaning of the original sentence by replacing {\em their idea} to {\em it}. 
Meanwhile, \proposed made the sentence more grammatical by adding a possessive {\em 's}.
The second example demonstrates challenging issues for future work in GEC. 
The correction by \mle looks fairly fluent as well as grammatical, but it is semantically nonsense. 
The correction by \proposed is also fairly fluent and makes sense, but the meaning has been changed too much.
For further improvement, better GEC models that are aware of the context or possess world knowledge are needed.


\section{Conclusions}
\label{sec:conclusions}
\vspace{-2mm}
We have presented a neural \encdec model with reinforcement learning for GEC. 
To alleviate the \mle issues (exposure bias and token-level optimization), \proposed learns the policy (model parameters) by directly optimizing toward the final objective by treating the final objective as the reward for the \encdec agent.
Using a GEC-specific metric, \metric, we have demonstrated that \proposed outperforms the \mle baseline on the fluency-oriented GEC corpus both in human and automated evaluation metrics.
As a supplement, we have explained the relevance between minimum risk training (\mrt) and \proposed, claiming that \mrt is a special case of \proposed.

\bibliography{emnlp2017}
\bibliographystyle{emnlp_natbib}

\newpage
\appendix
\section{Minimum Risk Training and Policy Gradient in Reinforcement Learning}
\setlength{\abovedisplayskip}{3.0pt} 
\setlength{\belowdisplayskip}{3.0pt} 
\label{sec:appendix}
We explain the relevance between minimum risk training (MRT) \cite{shen-EtAl:2016:P16-1} and neural reinforcement learning (\proposed) for training neural \encdec models.
We describe the detailed derivation of gradient in \mrt, and show that \mrt is a special case of \proposed.

As introduced in \S \ref{sec:model}, the model takes ungrammatical source sentences $x \in X$ as an input, and predicts grammatical and fluent output sentences $y \in Y$.
The objective function in \proposed and \mrt are written as follows.
\begin{align}
\label{eq:j}
J(\theta)= \mathbb{E}[\nrlreward]
\end{align}
\begin{align}
\label{eq:partialr}
R(\theta) =\sum_{\trainingdata} \mathbb{E}[\mrtdelta]
\end{align}
where $\nrlreward$ is the {\em reward} and $\mrtdelta$ is the {\em risk} for an output ($\hat{y}$).

To the sake of simplicity, we consider expected loss in \mrt for a single training pair:
\begin{align}
\tilde{R}(\theta) &= \mathbb{E}[\mrtdelta] \nonumber \\
    &= \sum_{\sampledata} \mrtq \mrtdelta
\end{align}
where
\begin{align}
\label{eq:softmax}
\mrtq = \softmax
\end{align}
$S(x)$ is a sampling function that produces $k$ samples $\hat{y}_1, ... \hat{y}_k$, and $\alpha$ is a smoothing parameter for the samples \cite{och:2003:ACL}.
Although the direction to optimize (i.e., minimizing or maximizing) is different, we see the similarity between $J(\theta)$ and $\tilde{R}(\theta)$ in the sense that they both optimize models directly towards evaluation metrics.

The partial derivative of $\tilde{R}(\theta)$ with respect to the model parameter $\theta$ is derived as follows. 
\begin{align}
\label{eq:partialrtilde}
\pderivrtilde &= \frac {\partial }{\partial \theta} \sum_{\sampledata} \mrtq \mrtdelta \nonumber \\
    &= \sum_{\sampledata} \mrtdelta \pderivtheta \mrtq
\end{align}

We need $\pderivtheta \mrtq$ in (\ref{eq:partialrtilde}).
For space efficiency, we use $\mrtqsimple$ as $\mrtq$ and $\mrtpsimple$ as $\mrtp$ below.

\begin{align}
\label{eq:partialq}
\pderivtheta \mrtqsimple &= \frac{\partial \mrtqsimple}{\partial \mrtpsimple} \frac{\partial \mrtpsimple}{\partial \theta} \ \ \ \ (\because \text{chain rule}) \nonumber \\
    &= \frac{\partial \mrtqsimple}{\partial \mrtpsimple} \nabla \mrtpsimple
\end{align}
For $\frac{\partial \mrtqsimple}{\partial \mrtpsimple}$, by applying the quotient rule to (\ref{eq:softmax}),
\begin{align}
\label{eq:qpartialp}
\frac{\partial \mrtqsimple}{\partial \mrtpsimple} &= \frac{\{\softmaxdenominator\} \pderivp \softmaxnumerator - \softmaxnumerator \pderivp \softmaxdenominator} {\{\softmaxdenominator\}{^2} } \nonumber \\
  &= \frac{\alpha \mrtpsimple ^{\alpha-1}} {\softmaxdenominator} - \frac{\alpha \mrtpsimple ^\alpha \mrtpsimple^{\alpha-1}} {\{\softmaxdenominator\}{^2}} \nonumber \\
  &= \alpha \frac{\mrtpsimple ^{\alpha-1}}{\softmaxdenominator} \left\{1- \frac{\mrtpsimple^\alpha}{\softmaxdenominator} \right\} \nonumber \\
  &= \alpha \frac{\mrtpsimple ^{\alpha}}{\softmaxdenominator} \frac{1}{\mrtpsimple} \left\{1- \frac{\mrtpsimple^\alpha}{\softmaxdenominator} \right\}
\end{align}
Thus, from (\ref{eq:partialq}) and (\ref{eq:qpartialp}), (\ref{eq:partialrtilde}) is
\begin{multline}
\pderivrtilde = \sum_{\sampledata} \mrtdelta \nabla \mrtpsimple \\ \left[ \alpha \frac{\mrtpsimple ^{\alpha}}{\softmaxdenominator} \frac{1}{\mrtpsimple} \left\{1- \frac{\mrtpsimple^\alpha}{\softmaxdenominator} \right\}  \right] \nonumber
\end{multline}
\begin{align}
\label{eq:partialr2}
  &= \alpha \mathbb{E} \left[\nabla\mrtpsimple \cdot \frac{1}{\mrtpsimple} \left\{ \mrtdelta - \mathbb{E}\left[\mrtdelta \right] \right\}  \right] \nonumber \\
  &= \alpha \mathbb{E} \left[\nabla \log \mrtpsimple \left\{ \mrtdelta - \mathbb{E}\left[\mrtdelta \right] \right\}  \right] 
\end{align}

According to the policy gradient theorem for REINFORCE \cite{williams1992simple,sutton1999policy}, the partial derivative of (\ref{eq:j}) is given as follows:
\begin{align}
\label{eq:partialj}
\pderivj = \tilde{\alpha} \mathbb{E} \left[\nabla \log \mrtpsimple \{\nrlreward - b \}  \right]
\end{align}
where $\tilde{\alpha}$ is a {\em learning rate}\footnote{In this appendix, we use $\tilde{\alpha}$ to distinguish it from smoothing parameter $\alpha$ in \mrt. } and $b$ is arbitrary {\em baseline reward} to reduce the variance of gradients.
Finally, we see that the gradient of \mrt (\ref{eq:partialr2}) is a special case of policy gradient in REINCOFCE (\ref{eq:partialj}) with $b = \mathbb{E}\left[\mrtdelta \right]$.
It is also interesting to see that the smoothing parameter $\alpha$ works as a part of learning rate ($\tilde{\alpha}$) in \proposed.

\end{document}